\documentclass{article}
\usepackage{spconf,amsmath,graphicx,amstext,amssymb,epsf}

\usepackage{epstopdf}

\title{Appearance Based Robot and Human Activity Recognition System}
\name{Bappaditya Mandal}
\address{Email: bmandal@i2r.a-star.edu.sg\\Institute for Infocomm Research, A*STAR, Singapore}
\begin{document}
%\ninept
%
\maketitle
\begin{abstract}
In this work, we present an appearance based human activity recognition system. It uses background modeling to segment the foreground object and extracts useful discriminative features for representing activities performed by humans and robots. Subspace based method like principal component analysis is used to extract low dimensional features from large voluminous activity images. These low dimensional features are then used to classify an activity. An apparatus is designed using a webcam, which watches a robot replicating a human fall under indoor environment. In this apparatus, a robot performs various activities (like walking, bending, moving arms) replicating humans, which also includes a sudden fall. Experimental results on robot performing various activities and standard human activity recognition databases show the efficacy of our proposed method.
\end{abstract}
\section{Introduction}
\label{sec:intro}

Automatic human activity recognition from video is an important problem that plays critical roles in many domains, such as health-care environments, surveillance, athletics and human-computer interactions  \cite{Sanchez,Mandal3,Leo1,Mandal2,Mandal7}. Developing algorithms to recognize human activities has proven to be an immense challenge since it is a problem that combines the uncertainty associated with computational vision with the added whimsy of human behavior. One of the fundamental challenges of recognizing activities is accounting for the variability that arises when cameras capture humans performing arbitrary actions \cite{Mandal5,Mandal4}. Popular survey papers for human activity recognition and their challenges can be found in \cite{Moeslund1,Moeslund2,Poppe1}.

In this work, an appearance based automatic activity recognition system is presented. This system captures images (videos) and recognizes the activities performed by humans/robots. Presently, this has been tested and evaluated on a pilot robot performing various activities, frontal and side poses in an open environment. Also, our algorithm is tested on publicly available human activity database. Our framework has five modules: 1. data acquisition (video) from the camera; 2. normalize the images; 3. extract features using subspace methods; 4. match the features with those of the stored templates in the database 5. output an activity recognition ID. We preprocess the images (incoming videos) using the normalization technique described in \cite{Beveridge,Mandal6,Jiang2} and then apply the popular statistical pattern recognition method principal component analysis (PCA) \cite{Fukunaga} to extract useful discriminating features for recognizing various activities.

\section{DESCRIPTION OF THE SYSTEM}
\label{sec:format}

\subsection{Apparatus Design}
\begin{figure}[!htp]
\centering
\includegraphics*[height=1.37in]{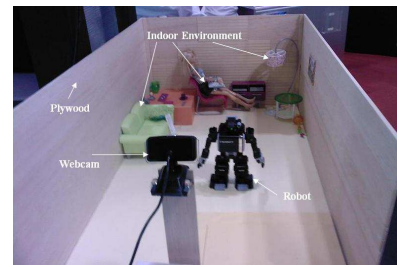}
\caption{The proposed robot activity recognition apparatus. It has indoor environment, robot walks freely inside this environment and being watched by a mounted webcam.}
\label{fig1_robotApparatus}
\end{figure}

Fig. \ref{fig1_robotApparatus} shows the design of our apparatus. We have tried to replicated a scenario in which a human being (in our case the robot) walks inside the room and performs some daily activities including sudden fall. Presently, our apparatus is made up of plywood and open on one side so that the robot can walk in normally inside the room. The top of the apparatus is open and this leads to uncontrolled lighting environment. The background is indoor environment, where we have some static indoor objects, like sofa, table and lamp. This whole room is being watched by a mounted webcam as shown in Fig. \ref{fig1_robotApparatus}. This webcam is connected (wired or wireless) to a PC. It captures the data continuously and transmits to the PC. The captured video is view independent since the robot performs various activities in any arbitrary directions within the view of the camera. Inside the PC, we have our modules running which processes these videos and output the activity state of the robot. In our case, the camera is static and can capture the videos during day and night.

\subsection{System Overview} \vspace{-0.4cm}
\begin{figure}[!htp]
\centering
\includegraphics*[width=3.4in]{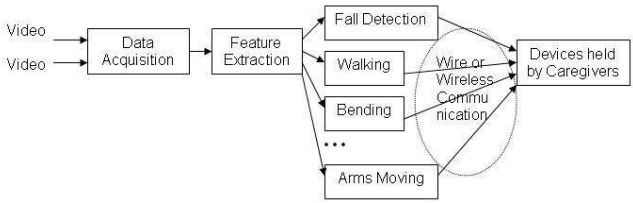}
\caption{Overview of our proposed system, which can detect and recognize four activities: arms moving, bending, falling and walking performed by a robot.}
\label{fig1_sysoverview}
\end{figure}
Fig. \ref{fig1_sysoverview} shows the system overview. Our first module is the image data (video) acquisition, which is performed using a simple webcam. The videos are captured and stored as image sequences. Currently our webcam captures images at 10 frames/sec with frame size $320\times240$. The proposed subspace based system framework involves two stages:\\
(i) \textit{Training:} This is a off-line stage, where some representative samples of the images are captured and used for training. During this stage, the machine learns the projection vectors where the training samples have maximum variance. The output of this stage is a set of basis vectors which captures most of the variance energy of the training samples and the projected templates of the training samples (in much reduced dimensions). Both the basis vectors and the templates of various activities are stored in the database.\\
(ii) \textit{Recognition:} This is a online process, where the system is tested with unseen images (videos). The new image samples are projected on the projection vectors obtained in (i) and matched against the stored templates in the database. The set of new sample images which matches closest to the stored temples is recognized as that recognition ID. Presently our system has recognition IDs like: Fall Detection, Walking, Bending and Arms Moving. For frontal and side poses estimation similar methodology is followed.

As soon as the recognition ID is found, if necessary, the system will generate an alert (SMS/email) to the Care-givers or nearby authorities. Presently, this system is working on a robot performing these activities in any arbitrary directions. We plan to extend it to many more other activities performed by human beings in indoor as well as outdoor environments.

\subsection{Foreground Silhouette Map Generation and Data Normalization}
\label{sec:Normalization}
\begin{figure}[!htp]
\centering
\begin{minipage}[b]{5.1in}
\includegraphics*[height=1.0in]{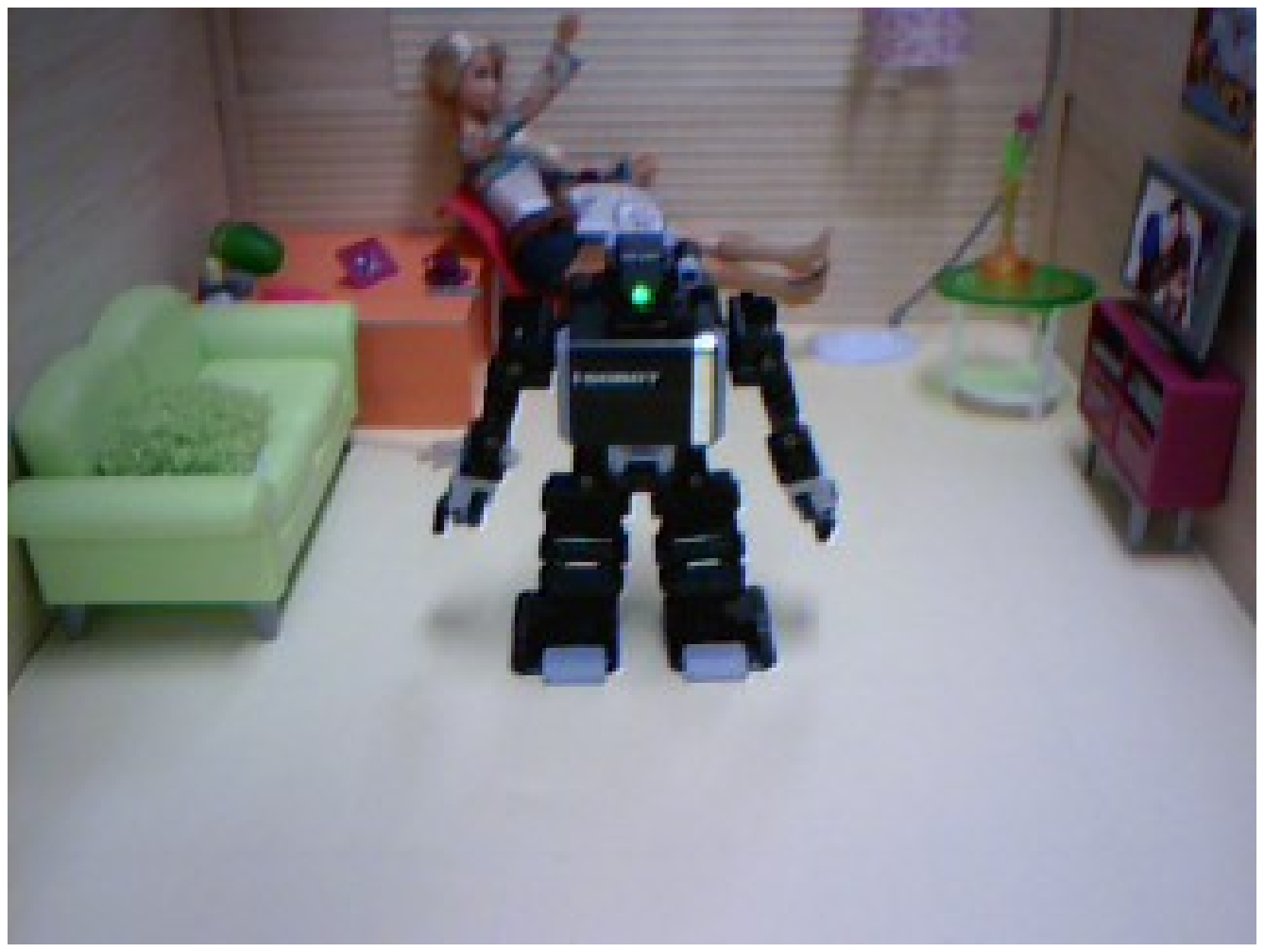}
\includegraphics*[height=1.0in]{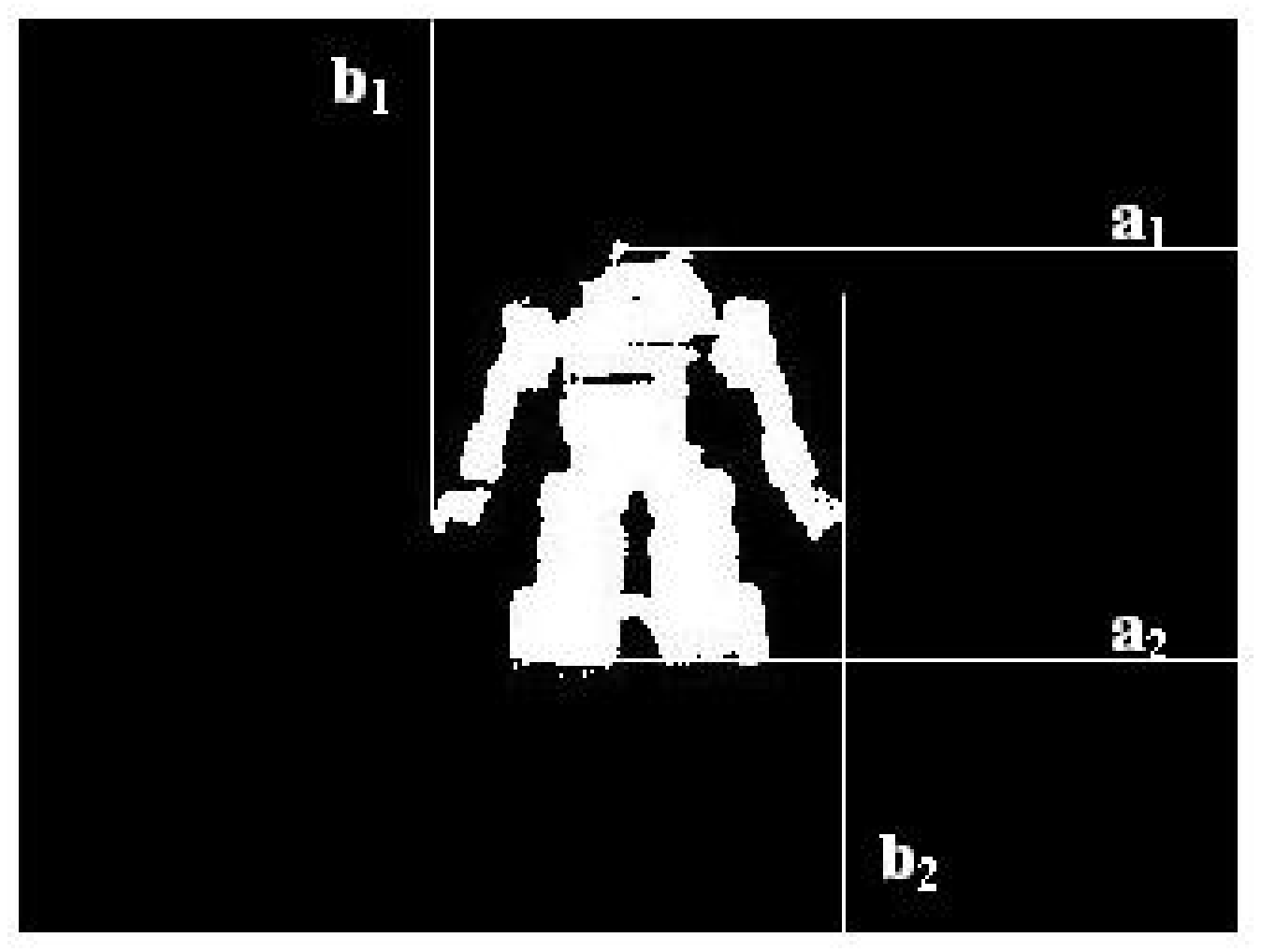}
\includegraphics*[height=0.7in]{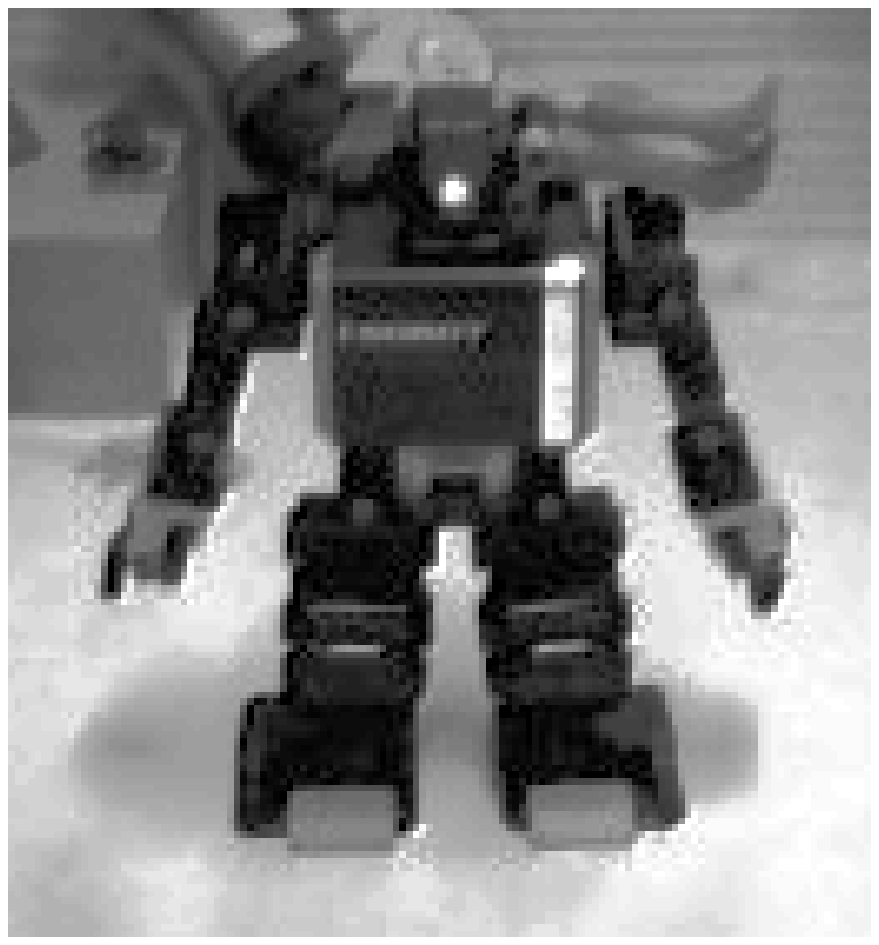}
\end{minipage}
\caption{Left: Original incoming image; Middle: silhouette image and centroid is calculated as $(a,b)=((a_1+a_2)/2,(b_1+b_2)/2)$; Right: normalized image. (Best viewed in color)}
\label{fig1_robot}
\end{figure}

The captured videos are processed automatically to detect the foreground from the background. We have adopted a method established in our previous work on background modeling and subtraction \cite{Lung1,Mandal9,Lu6}. The data captured are processed and divided into two types (a) color images of the scene, i.e. robot walking inside the indoor (room) environment as shown in Fig. \ref{fig1_robot} (left) and (b) binary image silhouettes of the robot as shown in Fig. \ref{fig1_robot} (middle). Using the silhouettes images, we estimate the centroid of the foreground - robot. $a_1$ and $a_2$ are the two extreme points obtained in the x-axis direction of the foreground object (robot). Similarly, $b_1$ and $b_2$ are the two extreme points in the y-axis direction. Centroid is obtained using the mean of these points. Fig. \ref{fig1_robot} (middle) shows the estimation of the points used for calculating centroid.
Using this centroid point, we crop an image size of $140 \times 130$ from the original color images. This cropped color image is then normalized by (i) converting it into a gray scale image, then (ii) histogram equalization is performed to smooth the distribution of grey values for all the pixels. A sample preprocessed image is shown in Fig. \ref{fig1_robot} (right), (iii) the image is normalized so that all pixels have mean zero and standard deviation one.

\subsection{Activity Training Using Subspace Based Method}
In this work, an \textit{action} denotes a short sequence of body configurations (arm still, body bending). It is usually, but not exclusively, defined by one or a few body parts \cite{Mandal5,Lu7}. An \textit{activity} denotes a sequence of body configurations over a longer span of time. Activities can be assembled from one or more actions and actions can specify details of an activity (e.g. falling with arms raised trying to hold a support).
Let the normalized images obtained be of size $w$-by-$h$, we can form a training set of column
vectors $\{X_{ij}\}$, where $X_{ij}\in \mathbb{R}^{n=wh}$ is called image vector, by
lexicographic ordering the pixel elements of image $j$ of activity $i$. Let the training set
contain $p$ activities and $q_i$ sample images for activity $i$. The number of total training sample
is $l=\sum_{i=1}^pq_i$. For activity recognition, each activity is a class with prior probability of
$c_i$. The total (mixture) scatter matrix $\mathbf{S}^t$ is defined by
\begin{equation}\label{eqn_1_St}
\mathbf{S}^t=\sum_{i=1}^{p}\frac{c_i}{q_i}\sum_{j=1}^{q_i}(X_{ij}-\overline{X})(X_{ij}-\overline{X})^T,
\end{equation}
where $\overline{X}=\sum_{i=1}^{p}\frac{c_i}{q_i}\sum_{j=1}^{q_i}X_{ij}$. If all classes have equal prior
probability, then $c_i=1/p$.

If we regard the elements of the image vector or the class mean vector as features, these preliminary features
will be de-correlated by solving the eigenvalue problem \cite{Duda}
\begin{equation}\label{eqn_2_Eigendecomp}
\mathbf{\Lambda}^t={\mathbf{\Phi}^t}^T\mathbf{S}^t\mathbf{\Phi}^t,
\end{equation}
where $\mathbf{\Phi}^t=[\phi^t_1,...,\phi^t_n]$ is the eigenvector matrix of $\mathbf{S}^t$, and
$\mathbf{\Lambda}^t$ is the diagonal matrix of eigenvalues $\lambda^t_1,...,\lambda^t_n$
corresponding to the eigenvectors. We assume that the eigenvectors are sorted according to the eigenvalues in descending order $\lambda^t_1\geq,...,\geq\lambda^t_n$. We perform the dimensionality reduction by selecting the principal projections/directions of the data with larger variances.

The first advantage is that we can represent each activity by low dimensional discriminative features \cite{Mandal1,Mandal2,Mandal3}. Secondly, the model parameters can be computed directly from the training data, for example, by diagonalizing the sample covariance matrix. So this system does not have any free parameter. This approach is less sensitive to the training data, number of samples per activity and noises present in the data.

\subsection{Feature Extraction and Activity Recognition}
After solving the eigenvalue problem in (\ref{eqn_2_Eigendecomp}), the dimensionality reduction is performed
here by keeping the eigenvectors with the $d$ largest eigenvalues $\mathbf{{\Phi}}_d^{t}=[{\phi^t_k}]_{k=1}^d=[{\phi^t_1},...,{\phi^t_d}],$ where $d$ is the number of features usually selected by a specific application. A set of projected features in the subspace $Y\in \mathbb{R}^d$ of any image $X$ can be obtained by representing training samples with new feature vectors, $Y={\mathbf{\Phi}^t_d}^TX.$\\
\textit{At the recognition stage:} Transform each $n$-D face image vector $X$ into $d$-D feature vector $Y$ by using the extraction matrix $\mathbf{\Phi^t}$ obtained in the training stage. Finally, apply a classifier trained on the gallery set to recognize the probe feature vectors.

Human/robot (depending on the timer) activities have inherent varying space-temporal structure. They vary if performed by different persons and even the same performer is not ever able to reproduce a movement exactly. So to compare two activities of different lengths. we use dynamic time warping (DTW) \cite{Boulgouris}, which performs a time alignment and normalization by computing a temporal transformation allowing two activities to be matched. An illustrative example with diagrams is shown in \cite{Boulgouris}. In all the experiments of this work, a simple first nearest neighborhood classifier (1-NNK) is applied. Euclidean distance measure is used to measure the distance between a probe feature vector and a gallery feature vector.

\vspace{-0.2cm}
\section{Experimental Results}
In this work we evaluate our proposed methodology on 3 datasets: (a) robot performing 4 activities, (b) estimation of robot pose for frontal and side views and (c) Weizmann dataset \cite{Blank1,Gorelick1} containing several actors performing 10 actions. (a) and (b) datasets are created by us and (c) is a publicly available dataset. We preprocess all the images following the normalization procedure described in section \ref{sec:Normalization}. Each dataset is partitioned into training and testing datasets. There is no overlap in the training and testing datasets. Moreover, in Weizmann dataset, there is no overlap in actors performing various activities in training and testing datasets.

\subsection{Results on Real Time Robot Activities}
\begin{figure}[!htp]
\centering
\begin{minipage}[b]{8.5cm}
\includegraphics*[width=0.8in]{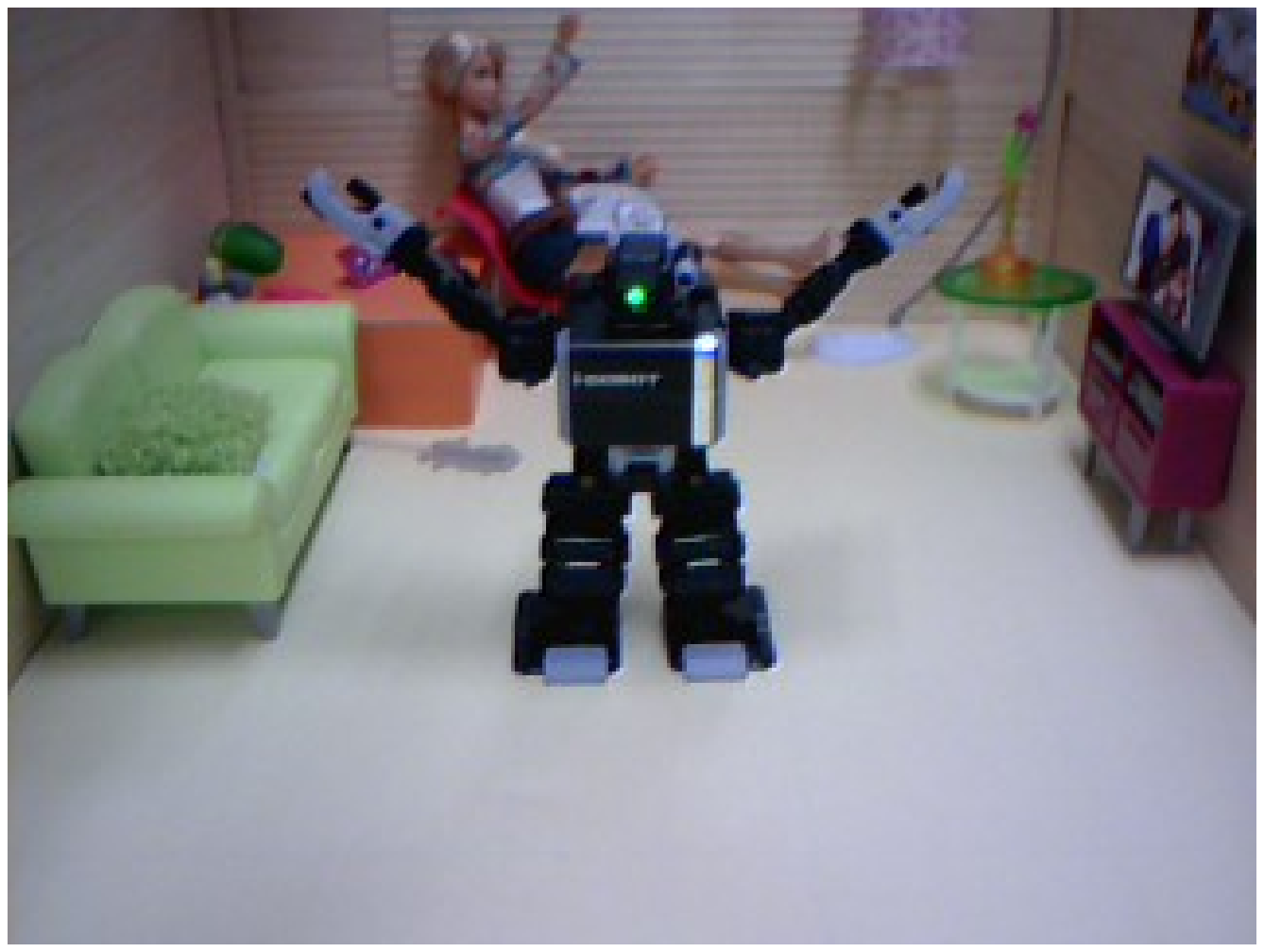}
\includegraphics*[width=0.8in]{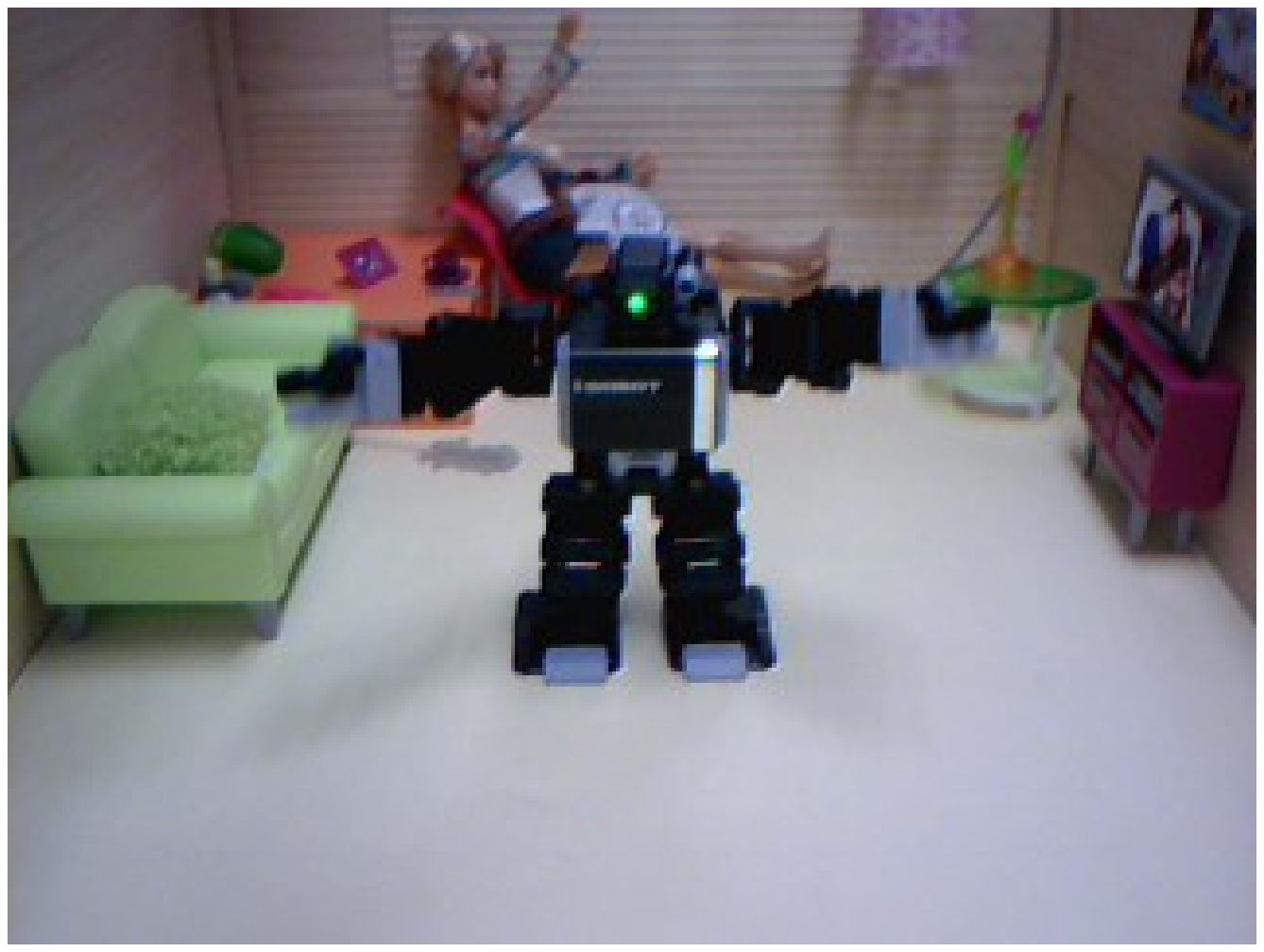}
\includegraphics*[width=0.8in]{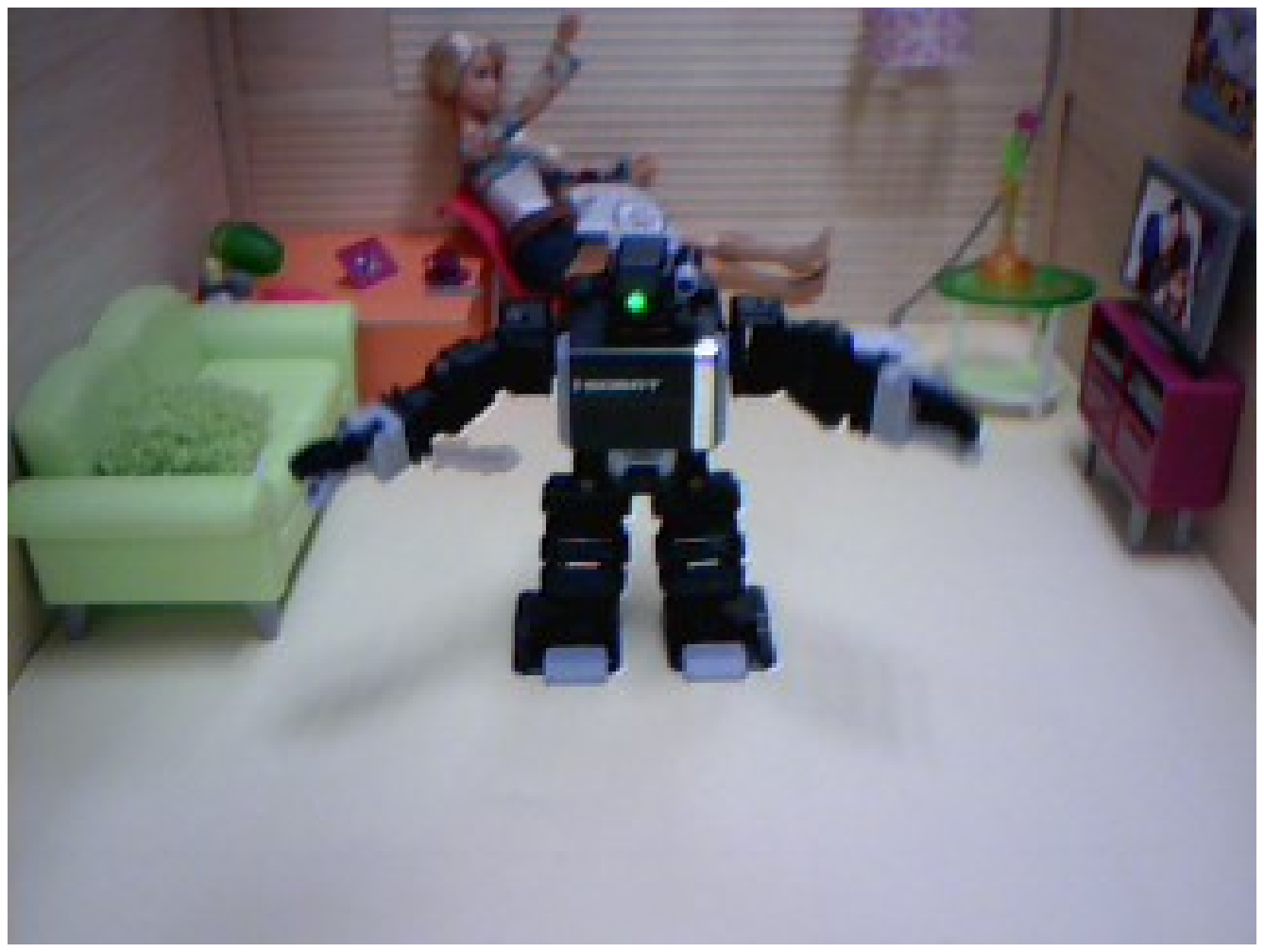}
\includegraphics*[width=0.8in]{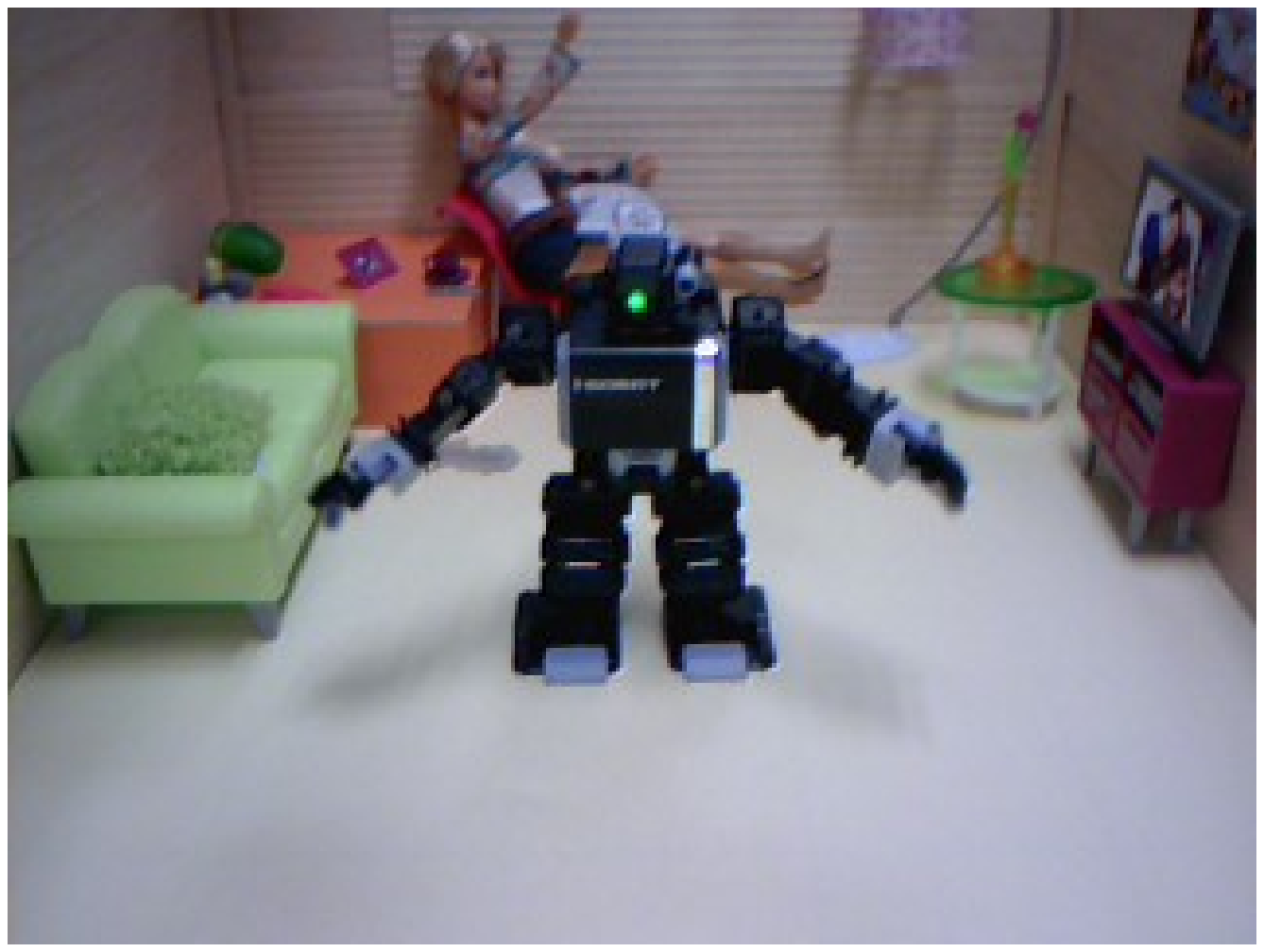}
\includegraphics*[width=0.8in]{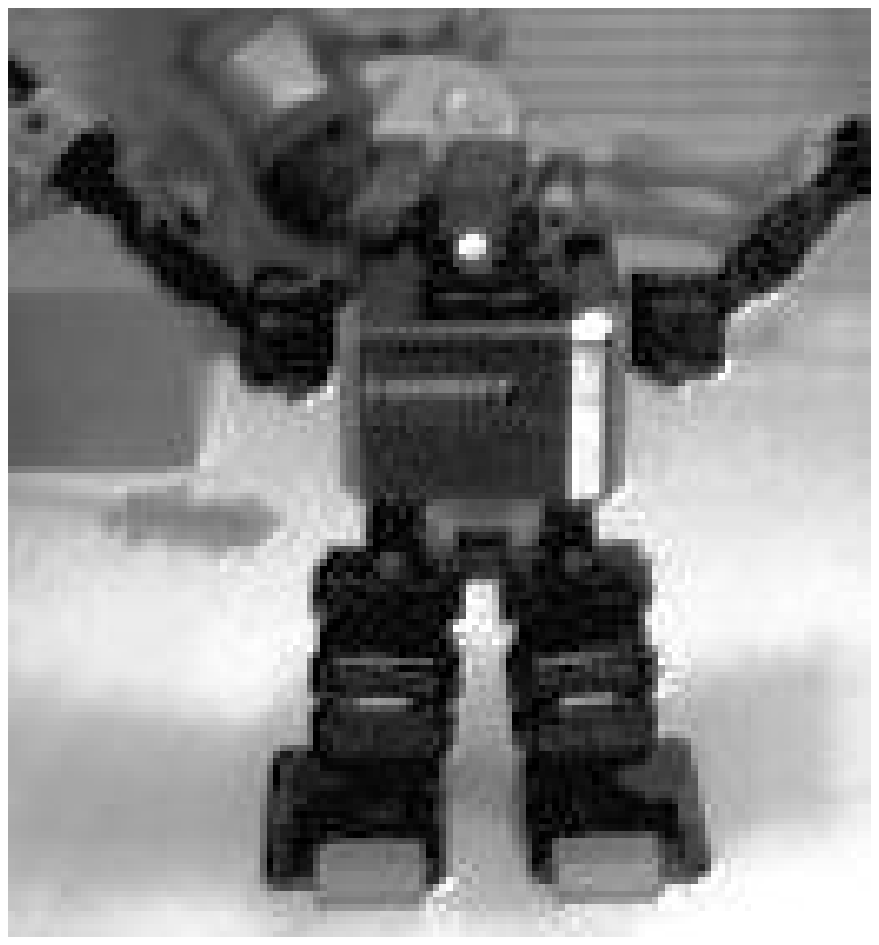}
\includegraphics*[width=0.8in]{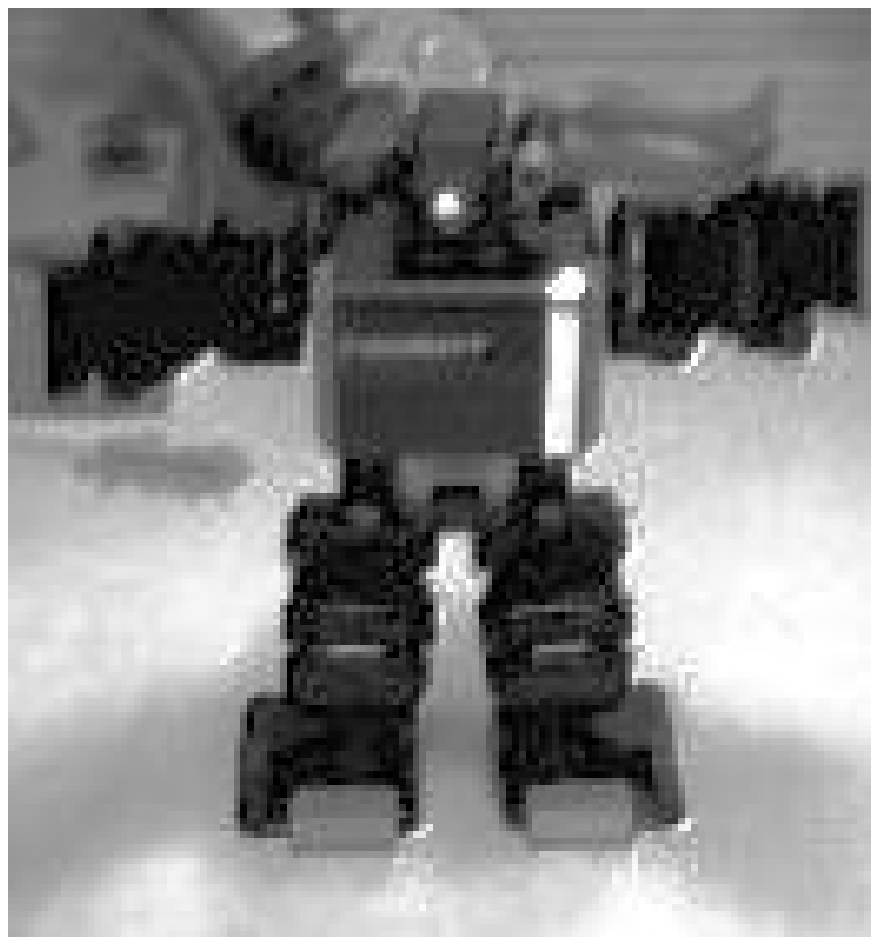}
\includegraphics*[width=0.8in]{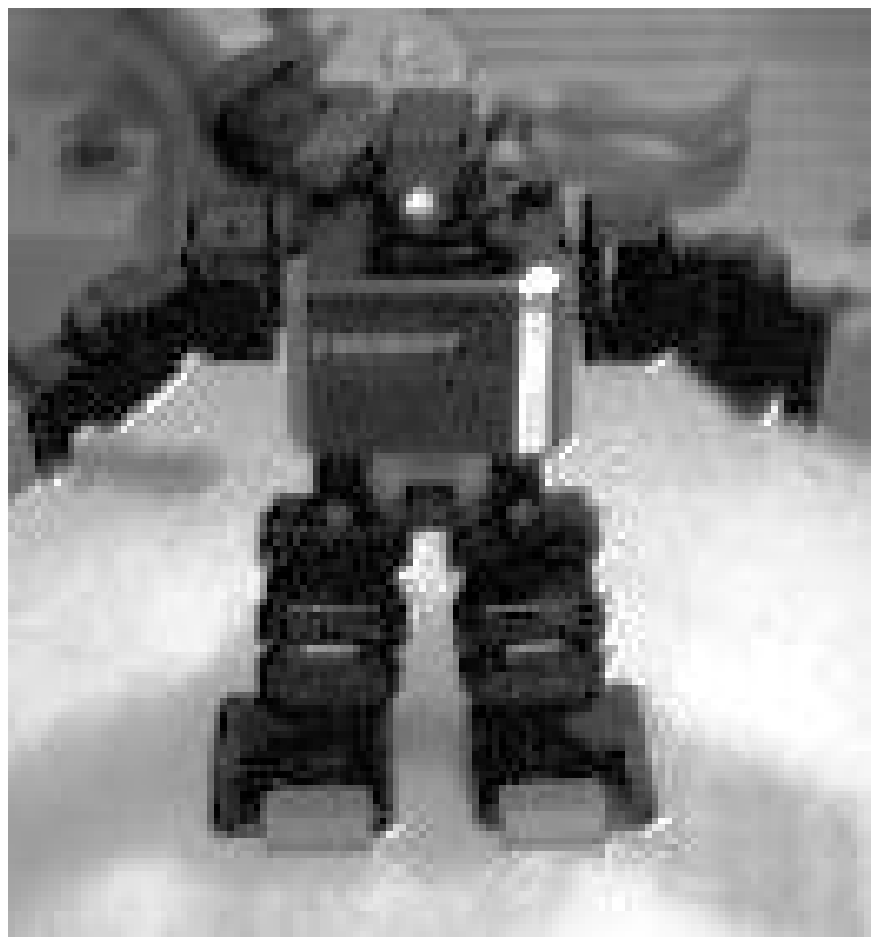}
\includegraphics*[width=0.8in]{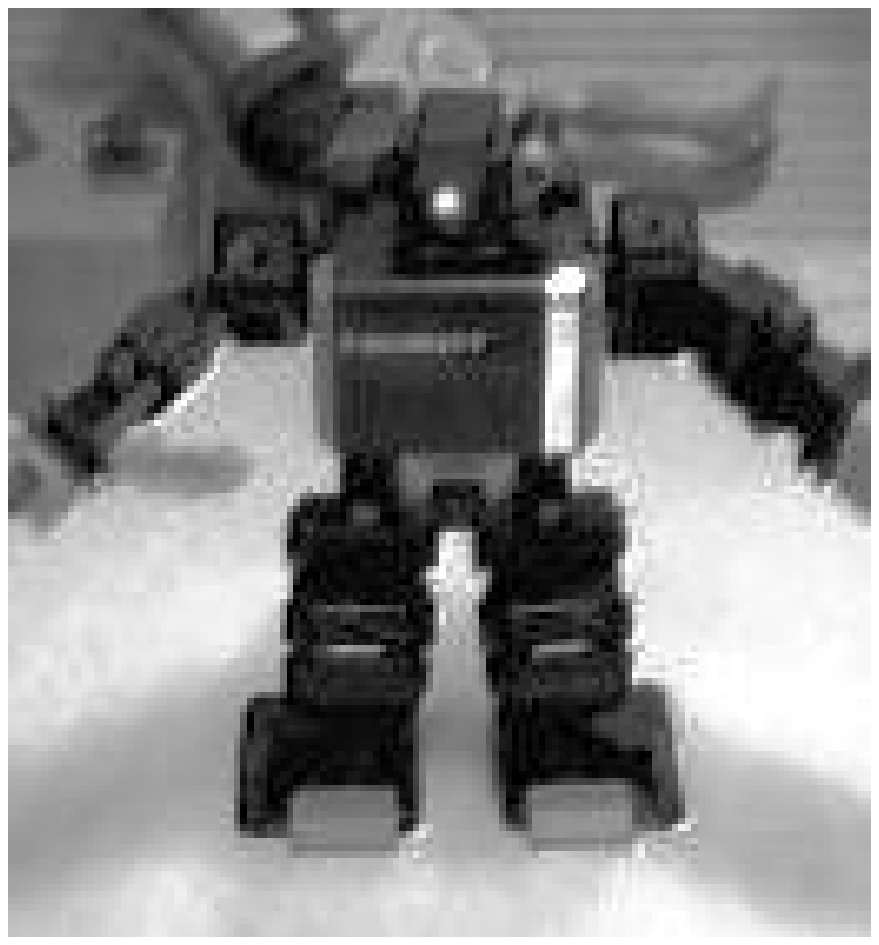}
\end{minipage} \\
\begin{minipage}[b]{8.5cm}
\includegraphics*[width=0.8in]{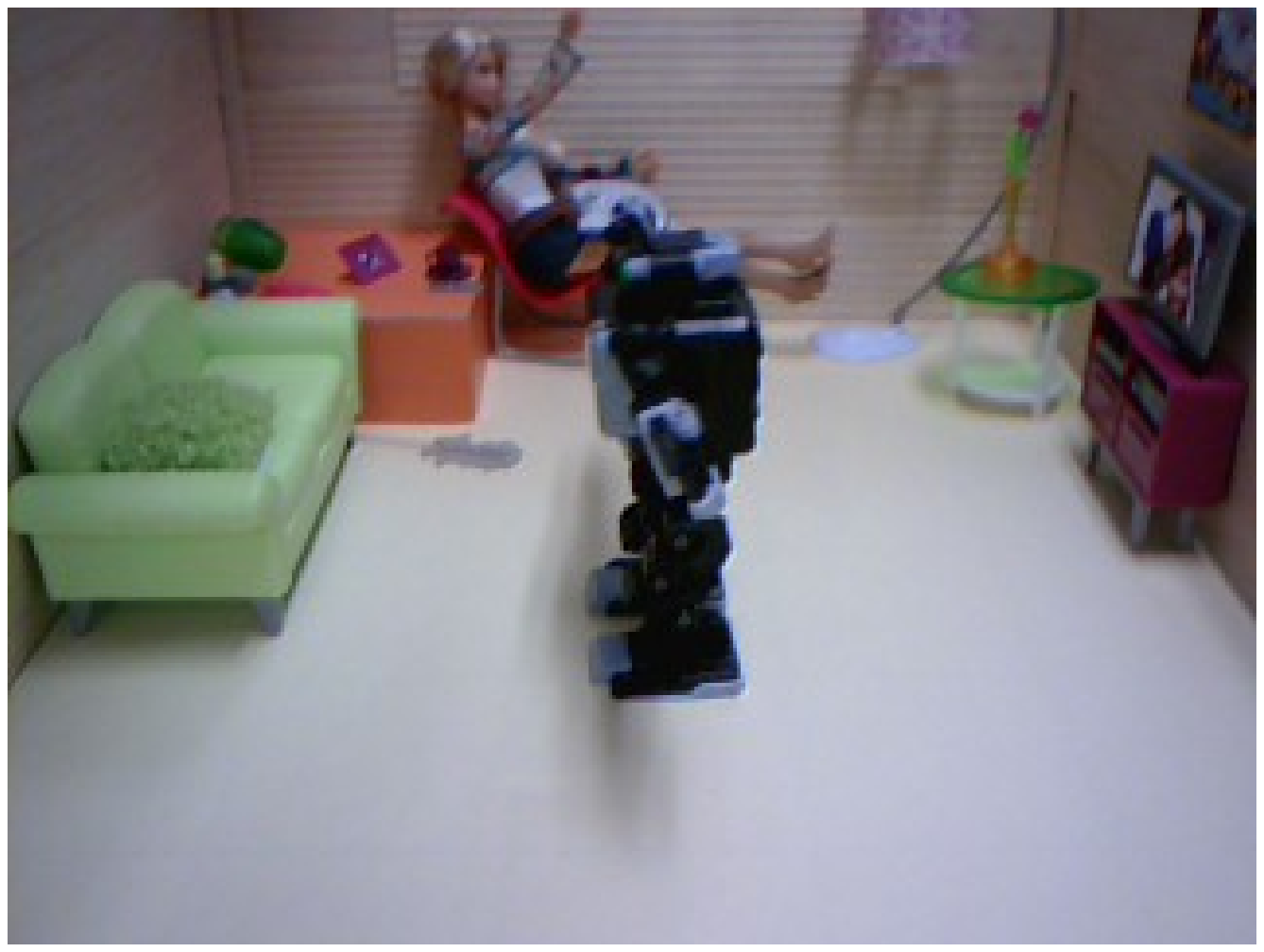}
\includegraphics*[width=0.8in]{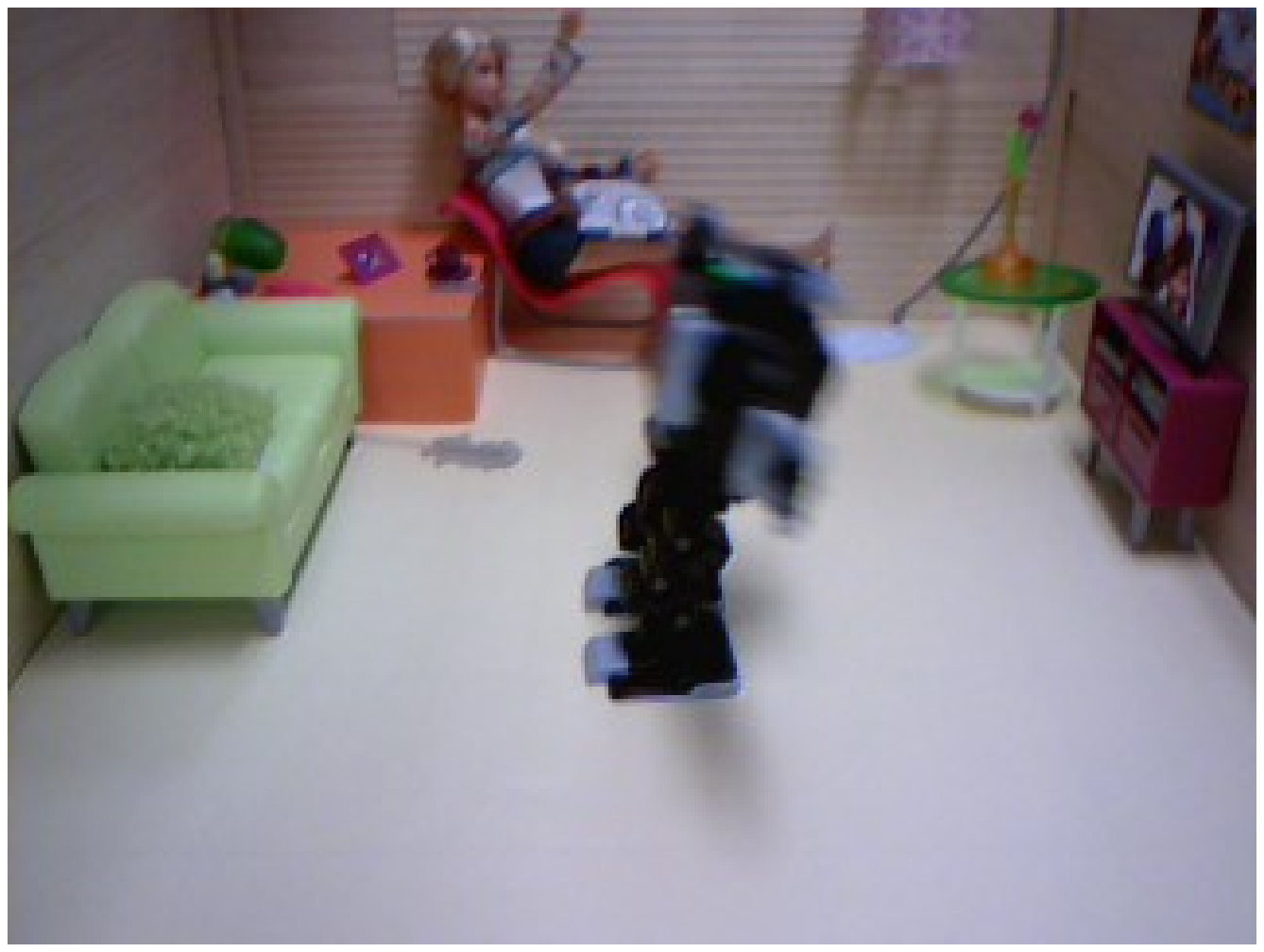}
\includegraphics*[width=0.8in]{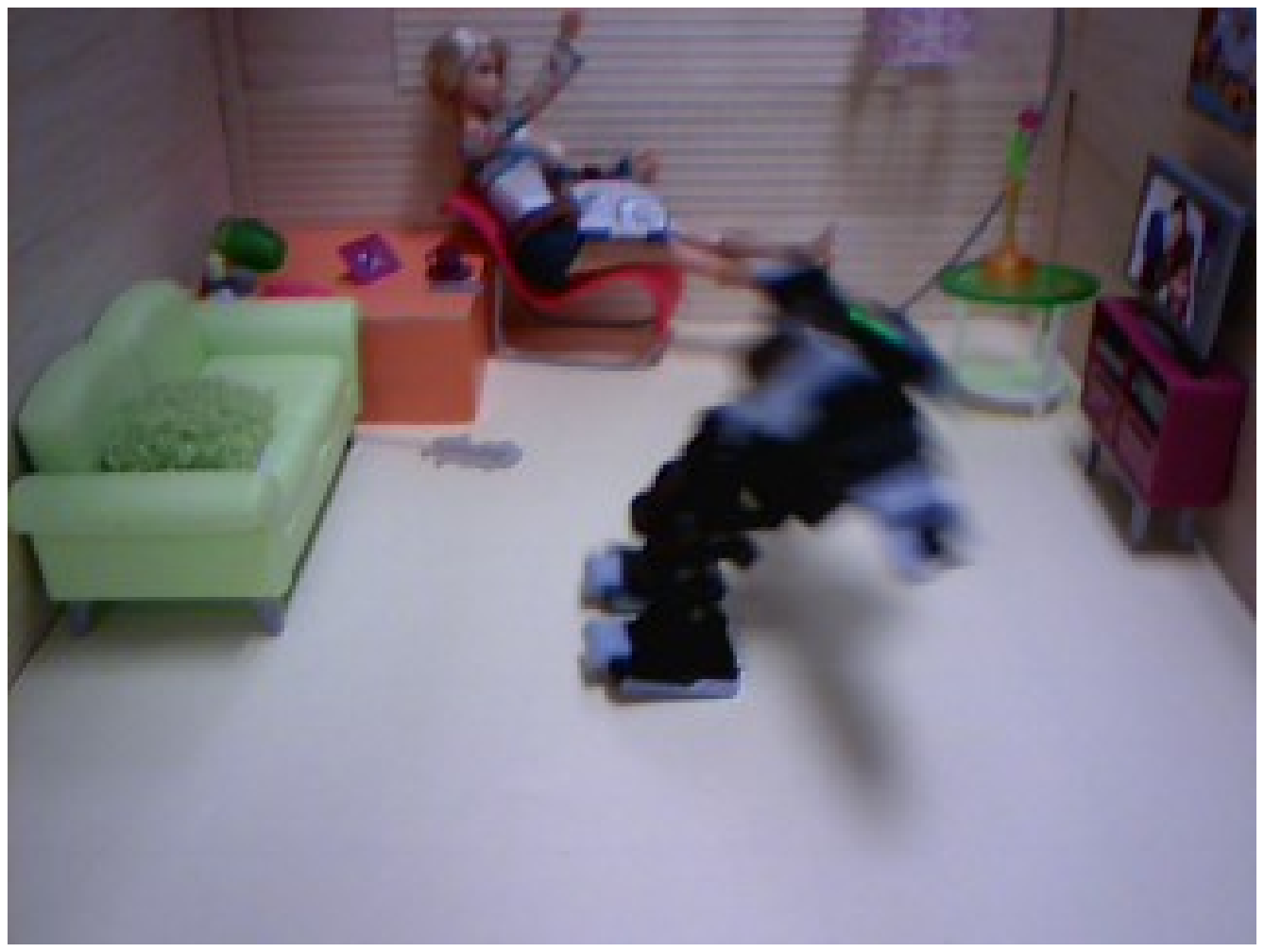}
\includegraphics*[width=0.8in]{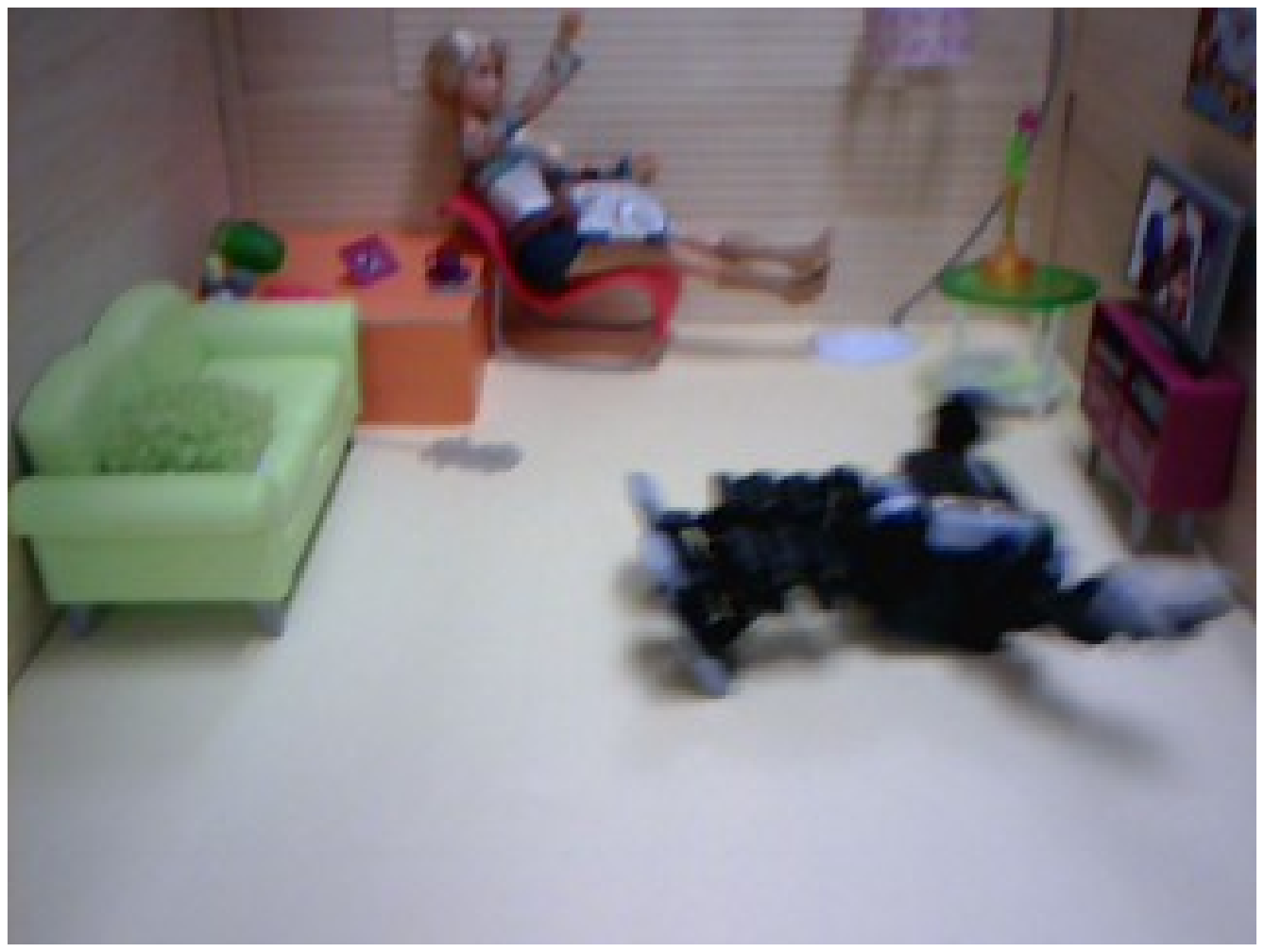}
\includegraphics*[width=0.8in]{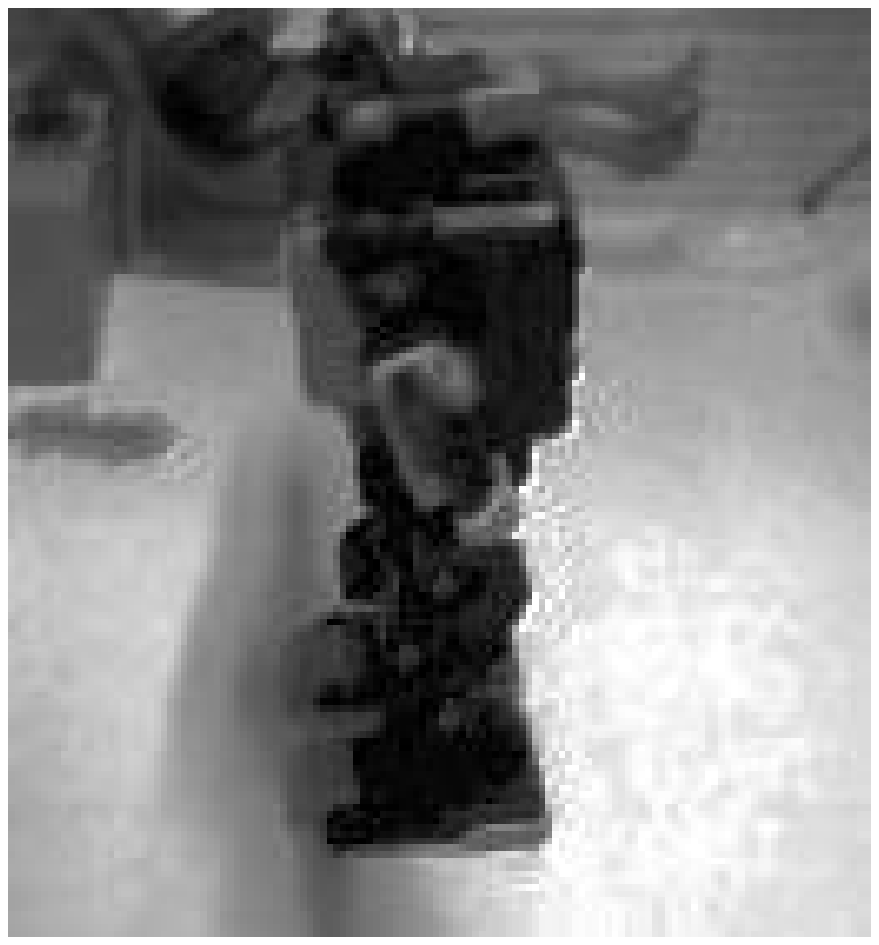}
\includegraphics*[width=0.8in]{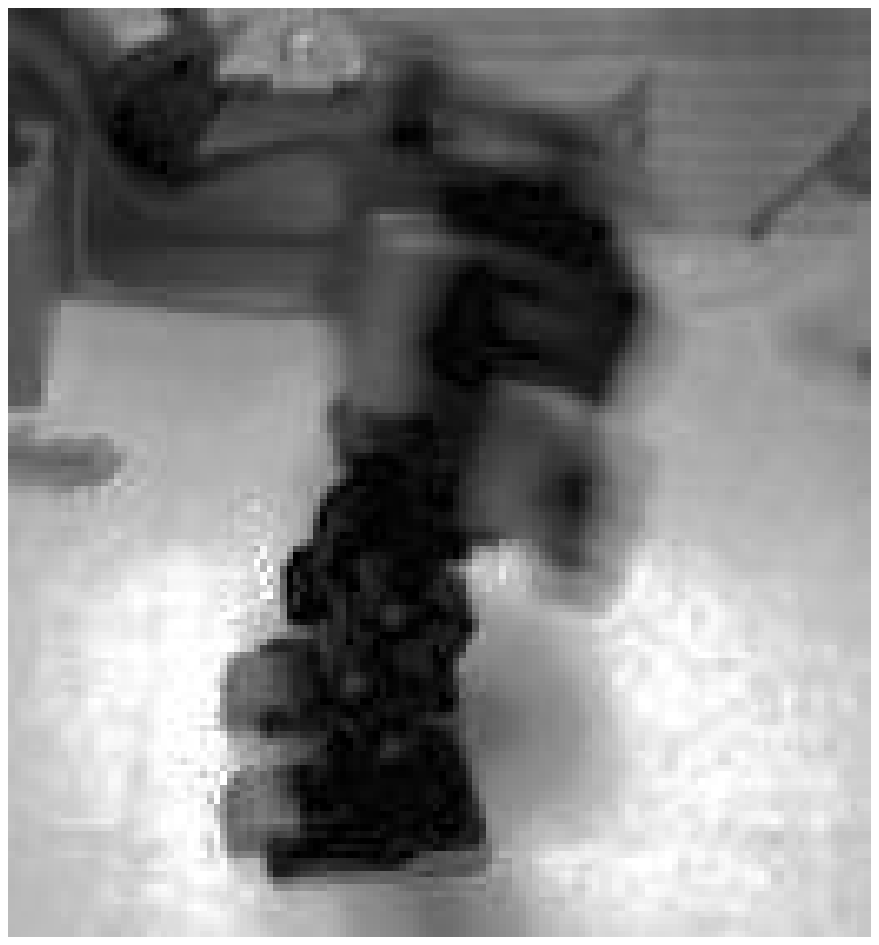}
\includegraphics*[width=0.8in]{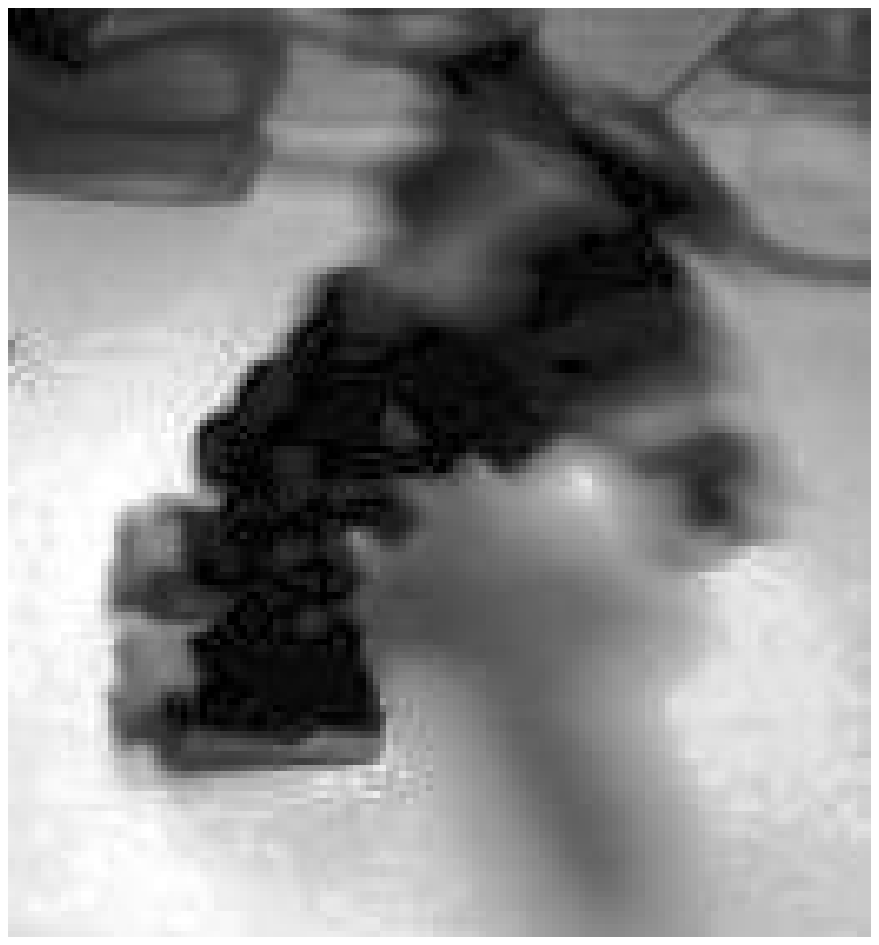}
\includegraphics*[width=0.8in]{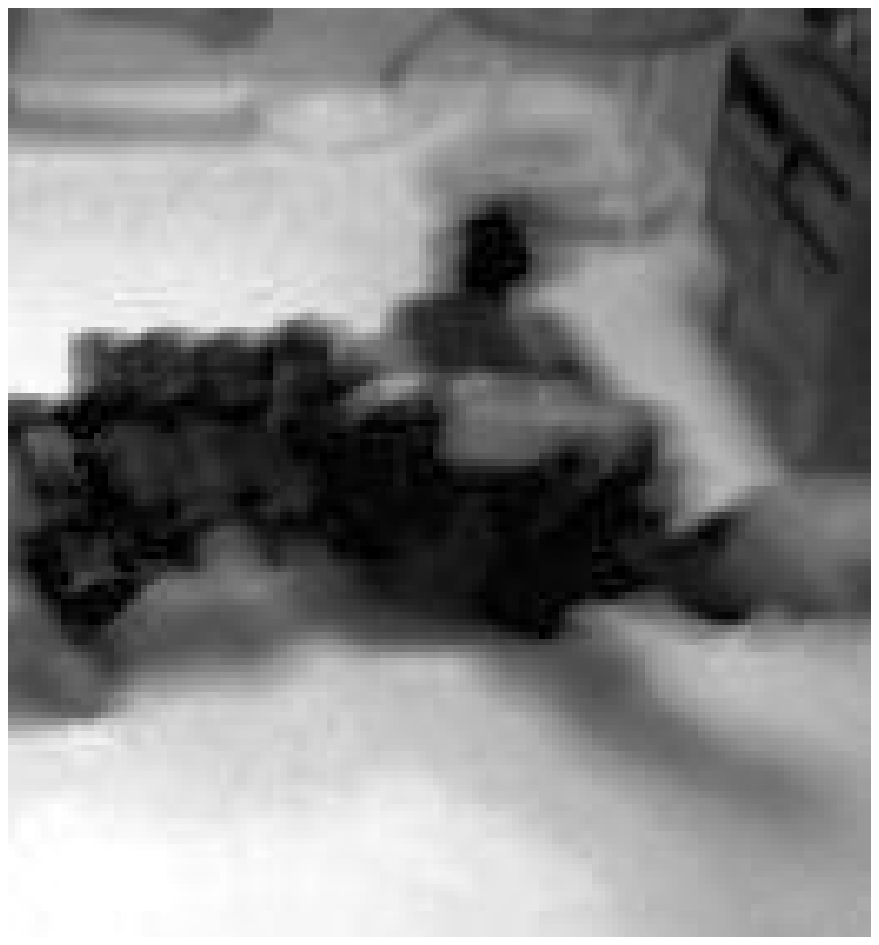}
\end{minipage}
\caption{Sample images from our Robot database for two activities (arms moving and falling) and their normalized images. Four sample images per activity. (Best viewed in color)}
\label{Fig_4_RobotNormImg}
\end{figure}

In our first experiment, we evaluate the proposed approach on a randomly picked up sample of video clip, which is done independently and at different times on pilot robot performing various activities: arms moving, bending, falling and walking. We divided the training and testing sessions and they are performed in the interval of 2 months. For training, a video clip of about 62 seconds (at 10 frames/sec) is captured and manually divided into four activities. Fig. \ref{Fig_4_RobotNormImg} shows the some sample robot images performing two activities: arms moving and falling. Their normalized images are also shown. Using  our webcam, independent testing video clip is obtained for 500 seconds. The video clip is captured at 10 frames/second, so 5000 testing images are obtained.

We created the ground truth for the testing sequences based on human knowledge of various activities and measured them against the activities recognized by our proposed system. Probe images captured in real time by our webcam in the interval of 10 frames/sec coming into the system are sampled for 1 sec and then our proposed methodology is applied. Table \ref{Tab_1} shows confusion matrix using the recognition rate (\%) for 5000 testing images (500 seconds of video clip). It is evident from the table that our system recognizes the falling (- an important aspect in heath-care environments) and bending very accurately. However, for arms moving and walking, our system performs poorly. This is probably because our system, presently, does not employ sophisticated spatiotemporal information. It uses only DTW to represent each class. Nevertheless, this system is presently working in real time and is able to detect very accurately the bending pose and falling - a scenario replication commonly encountered in elderly person falling. Moreover, the robot does bending many times while performing arms moving and walking. In this setup, we take only the first nearest neighborhood classifier (1-NNK). We anticipate that 2-NNK (first and second neighborhoods) would improve the results \cite{Mandal8}. E.g. $Bending-walking$ or $Bedning-arms~moving$ scenarios.
\begin{table}[!htp]
\caption{Confusion matrix of recognition rate (\%) using 5000 samples (video clip of 500 seconds) for four activities using 150 features. The rows represent the probe classes and columns represent the gallery classes} \label{Tab_1} \centering \footnotesize
\begin{tabular}{|c|c|c|c|c|c|c|c|c|c|c|c|}
\hline
Activities & $Arms$ & $Bending$ & $Falling$ & $Walking$\\
& $Moving$ & & &\\
\hline
$Arms~Moving$ & $8.04$ & $91.07$ & $0$ & $0.89$\\
\hline
$Bending$ & $0$ & $99.47$ & $0$ & $0.53$\\
\hline
$Falling$ & $1.96$ & $0$ & $93.14$ & $4.90$\\
\hline
$Walking$ & $0$ & $86.60$ & $0$ & $13.40$\\
\hline
\end{tabular}
\end{table}

\subsection{Results on Real Time Robot Poses}
In the second experiment, for training, we use 32 images divided equally for frontal and side poses. For testing, we use the same test sequence (of 5000 images) as used in the previous section and same experimental setup. However, this time we try to measure the frontal and side poses of the robot while it performs various activities. We perform this experiment so as to evaluate our algorithm for pose estimation. We anticipate that this estimation would help us in future algorithm development for real-time systems. Table \ref{Tab_2} shows confusion matrix using the recognition rate (\%) for 5000 testing images (500 seconds of video clip).
\begin{table}[!htp]
\caption{Confusion matrix of recognition rate (\%) using 5000 samples (video clip of 500 seconds) for frontal and side poses using 30 features. The rows represent the probe classes and columns represent the gallery classes} \label{Tab_2} \centering \small
\begin{tabular}{|c|c|c|c|c|c|c|c|c|c|c|c|}
\hline
Poses & $Frontal$ & $Side$\\
\hline
$Frontal$ & $57.93$ & $42.07$\\
\hline
$side$ & $9.17$ & $90.83$\\
\hline
\end{tabular}
\end{table}
From Table \ref{Tab_2} it is evident that our proposed algorithm estimated the side pose better than the frontal pose. More classes, like half left profile, half right profile would further help in getting more accurate pose estimations. We intend to develop these in our future work.

\subsection{Results on Weizmann dataset}
We have also evaluated the proposed approach on publicly available Weizmann dataset \cite{Blank1,Gorelick1}. It contains 10 actions: bend (bend), jumping-jack (jack), jump-in-place (pjump), jump-forward (jump), run (run), gallop-sideways (side), jump-forward-one-leg (skip), walk (walk), wave one hand (wave1), wave two hands (wave2), performed by 9 actors. Each of the video sequences ranges from 30 to 120 video frames. Silhouettes extracted from backgrounds and original image sequences are provided. We divide this database into training and testing sets. 4 actors performing 10 activities are used in the training while remaining 5 actors performing 10 activities are used in the testing. We run the experiments starting from 20 features to 200 features in the interval of 20 features. The results seem to be stable with 80 features, after which no improvement is obtained in recognition performance.
\begin{table}[!htp]
\caption{Confusion matrix of the recognition rates (\%) using our approach for 10 activities: (1) bend, (2) jack, (3) jump, (4) pjump, (5) run, (6) side, (7) skip, (8) walk, (9) wave1 and (10) wave2, averaged over testing samples of 5 performers.} \label{Tab_ActivityScores}
\centering
{\scriptsize
\begin{tabular}{|@{ }c@{ }|@{ }c@{ }|@{ }c@{ }|@{ }c@{ }|@{ }c@{ }|@{ }c@{ }|@{ }c@{ }|@{ }c@{ }|@{ }c@{ }|@{ }c@{ }|@{ }c@{ }|@{ }c@{ }|@{ }c@{ }|}
\hline
Activities & $1$ & $2$ & $3$ & $4$ & $5$ & $6$ & $7$ & $8$ & $9$ & $10$\\
\hline
1 & $96.67$ & $2.00$ & $1.33$ & $0$ & $0$ & $0$ & $0$ & $0$ & $0$ & $0$\\
\hline
2 & $2.33$ & $97.67$ & $0$ & $0$ & $0$ & $0$ & $0$ & $0$ & $0$ & $0$\\
\hline
3 & $0$ & $0$ & $96.67$ & $3.33$ & $0$ & $0$ & $0$ & $0$ & $0$ & $0$\\
\hline
4 & $0$ & $0$ & $2.67$ & $92.33$ & $0$ & $0$ & $3.33$ & $1.67$ & $0$ & $0$\\
\hline
5 & $0$ & $0$ & $0$ & $0$ & $95.33$ & $0$ & $0$ & $4.67$ & $0$ & $0$\\
\hline
6 & $0$ & $0$ & $0$ & $2.00$ & $0$ & $92.33$ & $5.67$ & $0$ & $0$ & $0$\\
\hline
7 & $0$ & $1.67$ & $0$ & $0$ & $0$ & $3.00$ & $95.33$ & $0$ & $0$ & $0$\\
\hline
8 & $0$ & $0$ & $0$ & $0$ & $3.67$ & $0$ & $0$ & $96.33$ & $0$ & $0$\\
\hline
9 & $0$ & $0$ & $0$ & $0$ & $0$ & $1.00$ & $0$ & $0$ & $95.33$ & $3.67$\\
\hline
10 & $0$ & $0$ & $0$ & $2.00$ & $0$ & $0$ & $0$ & $0$ & $1.33$ & $96.67$\\
\hline
\end{tabular}
}
\end{table}
Table \ref{Tab_ActivityScores} shows the confusion matrix using the activity recognition rate between the gallery and testing samples averaged over 5 performers for 10 activities. It is evident from the table that our system works well for most of the activities except pjump and side, where they are confused with skip. This is probably because they require more spatiotemporal information for better discrimination. We plan to incorporate this in our future algorithm developments.

\subsection{Summary and Conclusions}
In this work, we have developed a subspace based activity recognition system. Subspace or appearance based method like PCA is very much stable in noisy environment and not so sensitive to the training samples. Moreover, it handles the out-liners very efficiently. Sample covariance matrix is computed from the training data and after eigen decomposition, eigenvectors corresponding to the largest eigenvalues are selected. All gallery and test samples are projected onto this reduced subspace and low dimensional discriminative features are extracted. Experimental results on three databases show promising results. Using this methodology, presently, a realtime system is working, which can recognize a robot performing four activities and estimates frontal and side poses. Moreover, this system is also working on a publicly available database, where it can recognize many activities performed by humans.

% BiBTeX files (here: refs). The IEEEbib.bst bibliography
\bibliographystyle{IEEEbib}
\bibliography{face-activity}

\end{document}